\pdfoutput=1
\documentclass{article}
\usepackage{hyperref}
\usepackage{graphicx}
\usepackage{subfigure}
\usepackage[utf8]{inputenc} 
\usepackage[T1]{fontenc}    
\usepackage{amsmath}

\usepackage{hyperref}       
\usepackage{url}            
\usepackage{booktabs}       
\usepackage{amsfonts}       
\usepackage{nicefrac}       
\usepackage{microtype}      
\usepackage{color}

\DeclareMathOperator{\Tr}{Tr}

\usepackage{multirow}
\usepackage{amsmath}

\PassOptionsToPackage{numbers, compress}{natbib}

\usepackage[final]{nips_2018}
\usepackage{authblk}

\author[1]{Daniil Polykovskiy}
\author[1]{Alexander Zhebrak}
\author[2]{Benjamin Sanchez-Lengeling}
\author[3]{Sergey Golovanov}
\author[3]{Oktai Tatanov}
\author[3]{Stanislav Belyaev}
\author[3]{Rauf Kurbanov}
\author[3]{Aleksey Artamonov}
\author[1]{Vladimir Aladinskiy}
\author[1]{Mark Veselov}
\author[1]{Artur Kadurin}
\author[4]{Simon Johansson}
\author[4]{Hongming Chen}
\author[1,3,5]{Sergey Nikolenko}
\author[6,7,8]{Al\'an Aspuru-Guzik}
\author[1]{Alex Zhavoronkov}

\affil[1]{Insilico Medicine Hong Kong Ltd, Pak Shek Kok, New Territories, Hong Kong}
\affil[2]{Chemistry and Chemical Biology Department, Harvard University, Cambridge, MA 02143 USA}
\affil[3]{Neuromation OU, Tallinn, 10111 Estonia}
\affil[4]{Hit discovery, Discovery Sciences, Biopharmaceutics R\&D, AstraZeneca Gothenburg, Sweden}
\affil[5]{National Research University Higher School of Economics,  St. Petersburg, 190008, Russia}
\affil[6]{Department of Chemistry and Department of Computer Science, University of Toronto, Toronto, Ontario M5S 3H6, Canada}
\affil[7]{Vector Institute for Artificial Intelligence, Toronto, Ontario M5S 1M1, Canada}
\affil[8]{Biologically-Inspired Solar Energy Program, Canadian Institute for Advanced Research (CIFAR), Toronto, Ontario M5S 1M1, Canada}

\title{Molecular Sets (MOSES): A Benchmarking Platform for Molecular Generation Models}

\usepackage{natbib}
\usepackage{graphicx}

\begin{document}

\maketitle

\begin{abstract}
    Generative models are becoming a tool of choice for exploring the molecular space. These models learn on a large training dataset and produce novel molecular structures with similar properties. Generated structures can be utilized for virtual screening or training semi-supervised predictive models in the downstream tasks. While there are plenty of generative models, it is unclear how to compare and rank them. In this work, we introduce a benchmarking platform called Molecular Sets (MOSES) to standardize training and comparison of molecular generative models. MOSES provides a training and testing datasets, and a set of metrics to evaluate the quality and diversity of generated structures. We have implemented and compared several molecular generation models and suggest to use our results as reference points for further advancements in generative chemistry research. The platform and source code are available at \url{https://github.com/molecularsets/moses}.
\end{abstract}
\section{Introduction}
The discovery of new molecules for drugs and materials can bring enormous societal and technological progress, potentially curing rare diseases and providing a pathway for personalized precision medicine \citep{Lee2018-np}. However, complete exploration of the huge space of potential chemicals is computationally intractable; it has been estimated that the number of pharmacologically-sensible molecules is in the order of $10^{23}$ to $10^{80}$ compounds \citep{Reymond2015-az,Kirkpatrick2004-sz}. Often, this search is constrained based on already discovered structures and desired qualities such as solubility or toxicity. There have been many approaches to exploring the chemical space {\it in silico} and {\it in vitro}, including high throughput screening, combinatorial libraries, and evolutionary algorithms \citep{Curtarolo2013-kh, Hu2009-kr,Le2016-lm,Pyzer-Knapp2015-vi}. Recent works demonstrated that machine learning methods can produce new small molecules \citep{zhavoronkov2019deep, merk2018novo, merk2018tuning, Polykovskiy2018-rf} and peptides \citep{grisoni2018designing} showing biological activity.

Over the last few years, advances in machine learning, and especially in deep learning, have driven the design of new computational systems for modeling increasingly complex phenomena. One approach that has been proven fruitful for modeling molecular data is deep generative models. Deep generative models have found applications in a wide range of settings, from generating synthetic images \citep{Karras2017-at} and natural language texts \citep{Yu2016-vf}, to the applications in biomedicine, including the design of DNA sequences \citep{Killoran2017-ql}, and aging research \citep{Zhavoronkov2019}. One important field of application for deep generative models lies in the inverse design of drug compounds \citep{Sanchez-Lengeling2018-xe} for a given functionality (solubility,  ease of synthesis, toxicity). Deep learning also found other applications in biomedicine \citep{Ching2018, Mamoshina2016}, including target identification \citep{dnn_target2018}, antibacterial drug discovery \citep{ivanenkov2019identification}, and drug repurposing \citep{vanhaelen2017design, Aliper16}.

\begin{figure*}[h]
        \begin{center}
        \centerline{\includegraphics[width=1\textwidth]{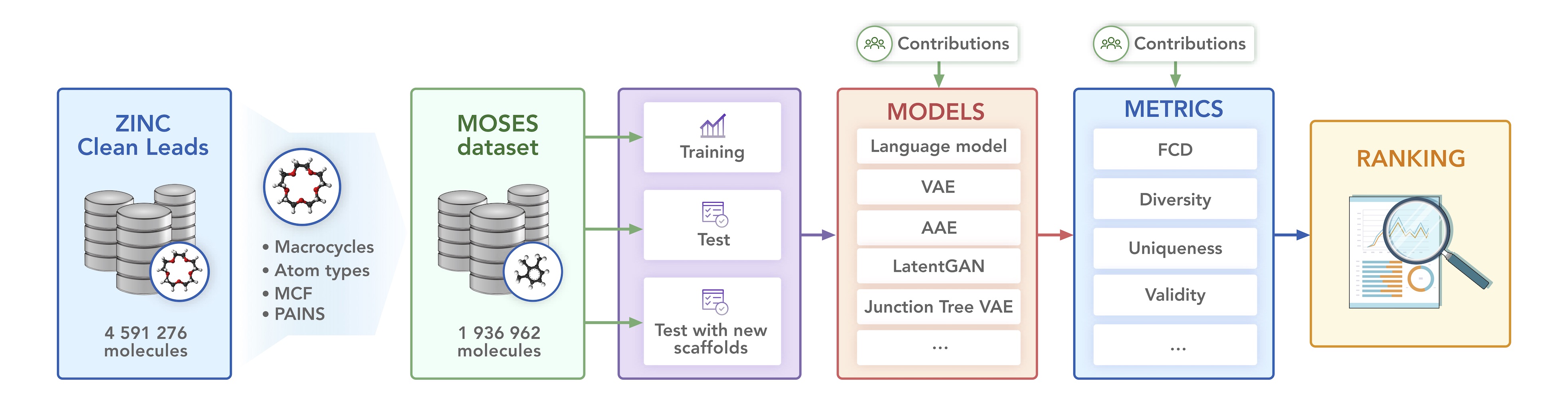}}
        \caption{Molecular Sets (MOSES) pipeline. The open-source library provides a dataset, baseline models, and evaluation metrics.}
        \label{fig:pipeline}
        \end{center}
\end{figure*}

Part of the success of deep learning in different fields has been driven by ever-growing availability of large datasets and standard benchmark sets. These sets serve as a common measuring stick for newly developed models and optimization strategies \citep{lecun1998gradient, Deng2009-xb}. In the context of organic molecules, MoleculeNet \citep{Wu2018-mr} was introduced as a standardized benchmark suite for regression and classification tasks. \citet{brown2019guacamol} proposed to evaluate generative models on goal-oriented and distribution learning tasks with a focus on the former. We focus on standardizing metrics and data for the distribution learning problem that we introduce below.

In this work, we provide a benchmark suite---Molecular Sets (MOSES)---for molecular generation: a standardized dataset, data preprocessing utilities, evaluation metrics, and molecular generation models. We hope that our platform will serve as a clear and unified testbed for current and future generative models. We illustrate the main components of MOSES in Figure~\ref{fig:pipeline}.

\section{Distribution learning}

In MOSES, we study distribution learning models. Formally, given a set of training samples $X_{\mathrm{tr}} = \left\{x^{\mathrm{tr}}_1, \dots, x^{\mathrm{tr}}_N\right\}$ from an unknown distribution $p(x)$, distribution learning models approximate $p(x)$ with some distribution $q(x)$. 

Distribution learning models are mainly used for building virtual libraries \citep{van2019virtual} for computer-assisted drug discovery. While imposing simple rule-based restrictions on a virtual library (such as maximum or minimum weight) is straightforward, it is unclear how to apply implicit or soft restrictions on the library. For example, a medicinal chemist might expect certain substructures to be more prevalent in generated structures. Relying on a set of manually or automatically selected compounds, distribution learning models produce a larger dataset, preserving implicit rules from the dataset. Another application of distribution learning models is extending the training set for downstream semi-supervised predictive tasks: one can add new unlabeled data by sampling compounds from a generative model.

The quality of a distribution learning model is a deviation measure between $p(x)$ and $q(x)$. The model can either define a probability mass function $q(x)$ implicitly or explicitly. Explicit models such as Hidden Markov Models, n-gram language models, or normalizing flows \citep{realnvp, shi2020graphaf} can analytically compute $q(x)$ and sample from it. Implicit models, such as variational autoencoders, adversarial autoencoders, or generative adversarial networks \citep{Gomez-Bombarelli2018-jp, Kadurin2017-rv, De_Cao2018-ju} can sample from $q(x)$, but can not compute the exact values of the probability mass function. To compare both kinds of models, evaluation metrics considered in this paper depend only on samples from $q(x)$.

\section{Molecular representations}
\begin{figure}[h]
        \begin{center}
        \centerline{\includegraphics[width=0.5\columnwidth]{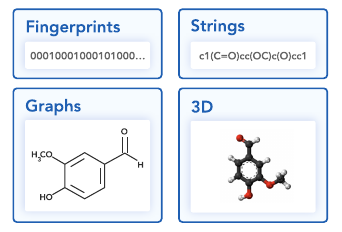}}
        \caption{Different views on a vanillin molecule.}
        \label{fig:representations}
        \end{center}
\end{figure}

In this section, we discuss different approaches to representing a molecule in a machine learning-friendly way (Figure~\ref{fig:representations}): string and graph representations.

{\bf String representations.} Representing a molecular structure as a string have been quickly adopted \citep{Popova2018-ut,Segler2018-eo,Jaques2016-dw,Guimaraes2017-ze,Olivecrona2017-vm,Kang2018-uy,Yang2017-jk,Kadurin2017-ki,Putin2018-sf} for generative models due to the abundance of sequence modeling tools such as recurrent neural networks, attention mechanisms, and dilated convolutions. Simplified molecular input line entry system (SMILES) \citep{Weininger1970} is the most widely used string representation for generative machine learning models. SMILES algorithm traverses a spanning tree of a molecular graph in depth-first order and stores atom and edge tokens. SMILES also uses special tokens for branching and edges not covered with a spanning tree. Note that since a molecule can have multiple spanning trees, different SMILES strings can represent a single molecule. While there is a canonicalization procedure to uniquely construct a SMILES string from a molecule \citep{Weininger1989}, ambiguity of SMILES can also serve as augmentation and improve generative models \citep{Arus-Pous2019-zg}.

DeepSMILES \citep{OBoyle2018-fj} was introduced as an extension of SMILES that seeks to reduce invalid sequences by altering syntax for branches and ring closures. Some methods try to incorporate SMILES syntax into a network architecture to increase the fraction of valid molecules \citep{Kusner2017-zj, Dai2018-dj}. SELFIES \citep{Krenn2019-dj} defines a new syntax based on a Chomsky type-2 grammar augmented with self-referencing functions. International Chemical Identifier (InChI) \citep{stein2003open} is a more verbose string representation which explicitly specifies a chemical formula, atoms' charges, hydrogens, and isotopes. However, \citet{Gomez-Bombarelli2018-jp} reported that InChI-based models perform substantially worse than SMILES-based models in generative modeling---presumably due to a more complex syntax.

{\bf Molecular graphs}. Graph representations have long been used in chemoinformatics for storing and processing molecular data. In a molecular graph, each node corresponds to an atom and each edge corresponds to a bond. Such graph can specify hydrogens either explicitly or implicitly. In the latter case, the number of hydrogens can be deduced from atoms' valencies.

Classical machine learning methods mostly utilize molecular descriptors extracted from such graphs. Deep learning models, however, can learn from graphs directly with models such as Graph Convolutional Networks \citep{Duvenaud2015-rn}, Weave Networks \citep{Wu2018-mr}, and Message Passing Networks \citep{Gilmer2017-vy}. Molecular graph can also be represented as adjacency matrix and node feature matrix; this approach has been successfully employed in the MolGAN model \citep{De_Cao2018-ju} for the QM9 dataset \citep{Ramakrishnan2014-ou}. Other approaches such as Junction Tree VAE \citep{Jin2018-sm} process molecules in terms of their subgraphs.

\section{Metrics} \label{sec:metrics}
In this section, we propose a set of metrics to assess the quality of generative models. The proposed metrics detect common issues in generative models such as overfitting, imbalance of frequent structures or mode collapse.
Each metric depends on a generated set $G$ and a test (reference) set $R$. We compute all metrics (except for validity) only for valid molecules from the generated set. We suggest generating $30{,}000$ molecules and obtaining $G$ as valid molecules from this set. 

{\bf Fraction of valid (Valid) and unique (Unique@k) molecules} report validity and uniqueness of the generated SMILES strings. We define validity using RDKit's molecular structure parser that checks atoms' valency and consistency of bonds in aromatic rings.  In the experiments, we compute Unique@$K$ and for the first $K=1{,}000$ and $K=10{,}000$ valid molecules in the generated set. If the number of valid molecules is less than $K$, we compute uniqueness on all valid molecules. Validity measures how well the model captures explicit chemical constraints such as proper valence. Uniqueness checks that the model does not collapse to producing only a few typical molecules.

{\bf Novelty} is the fraction of the generated molecules that are not present in the training set. Low novelty indicates overfitting.

{\bf Filters} is the fraction of generated molecules that pass filters applied during dataset construction (see Section \ref{sec:dataset}). While the generated molecules are often chemically valid, they may contain unwanted fragments: when constructing the training dataset, we removed molecules with such fragments and expect the models to avoid producing them.

{\bf Fragment similarity (Frag)} compares distributions of BRICS fragments \citep{Degen2008-fk} in generated and reference sets. Denoting $c_f(A)$ a number of times a substructure $f$ appears in molecules from set $A$, and a set of fragments that appear in either $G$ or $R$ as $F$, the metric is defined as a cosine similarity:
\begin{equation}
    \textrm{Frag}(G, R) = \frac{\sum\limits_{f \in F} \Big( c_f(G) \cdot c_f(R) \Big)}{\sqrt{\sum\limits_{f \in F} c^2_f(G)}\sqrt{\sum\limits_{f \in F} c^2_f(R)}}.
\end{equation}
If molecules in both sets have similar fragments, Frag metric is large. If some fragments are over- or underrepresented (or never appear) in the generated set, the metric will be lower. Limits of this metric are $[0, 1]$.

{\bf Scaffold similarity (Scaff)} is similar to fragment similarity metric, but instead of fragments we compare frequencies of Bemis–Murcko scaffolds \citep{Bemis1996-js}. Bemis–Murcko scaffold contains all molecule's ring structures and linker fragments connecting rings. We use RDKit implementation of this algorithm which additionally  considers carbonyl groups attached to rings as part of a scaffold. Denoting $c_s(A)$ a number of times a scaffold $s$ appears in molecules from set $A$, and a set of fragments that appear in either $G$ or $R$ as $S$, the metric is defined as a cosine similarity:
\begin{equation}
    \textrm{Frag}(G, R) = \frac{\sum\limits_{s \in S} \Big( c_s(G) \cdot c_s(R) \Big)}{\sqrt{\sum\limits_{s \in S} c^2_s(G)}\sqrt{\sum\limits_{s \in S} c^2_s(R)}}.
\end{equation}
The purpose of this metric is to show how similar are the scaffolds present in generated and reference datasets. For example, if the model rarely produces a certain chemotype from a reference set, the metric will be low. Limits of this metric are $[0, 1]$.

Note that both fragment and scaffold similarities compare molecules at a substructure level. Hence, it is possible to have a similarity $1$ even when $G$ and $R$ contain different molecules. 

{\bf Similarity to a nearest neighbor (SNN)} is an average Tanimoto similarity $T(m_G, m_R)$ (also known as the Jaccard index) between fingerprints of a molecule $m_G$ from the generated set $G$ and its nearest neighbor molecule $m_R$ in the reference dataset $R$:
\begin{equation}
    \textrm{SNN}(G, R) = \frac{1}{|G|}\sum_{m_G \in G}\max_{m_R \in R}T(m_G, m_R),
\end{equation}
In this work, we used standard Morgan (extended connectivity) fingerprints \citep{rogers2010extended} with radius $2$ and $1024$ bits computed using RDKit library \citep{Landrum_undated-rr}. The resulting similarity metric can be interpreted as precision: if generated molecules are far from the manifold of the reference set, similarity to the nearest neighbor will be low. Limits of this metric are $[0, 1]$.

{\bf Internal diversity ($\textrm{IntDiv}_p$)} \citep{Benhenda2017-yt} assesses the chemical diversity within the generated set of molecules $G$.
\begin{equation}
    \textrm{IntDiv}_p(G) =1 - \sqrt[p]{\frac{1}{|G|^2}\sum_{m_1, m_2 \in G}T(m_1, m_2)^p}.
\end{equation}
This metric detects a common failure case of generative models---mode collapse. With mode collapse, the model produces a limited variety of samples, ignoring some areas of the chemical space. A higher value of this metric corresponds to higher diversity in the generated set. In the experiments, we report $\textrm{IntDiv}_1(G)$ and $\textrm{IntDiv}_2(G)$. Limits of this metric are $[0, 1]$.

{\bf Fr\'echet ChemNet Distance (FCD)} \citep{Preuer2018-pf} is calculated using activations of the penultimate layer of a deep neural network ChemNet trained to predict biological activities of drugs. We compute activations for canonical SMILES representations of molecules.  These activations capture both chemical and biological properties of the compounds. For two sets of molecules $G$ and $R$, FCD is defined as
\begin{equation}
\textrm{FCD}(G, R) = \lVert \mu_G-\mu_R \rVert^2+\Tr\left(\Sigma_G+\Sigma_R-2(\Sigma_G\Sigma_R)^{1/2}\right)
\end{equation}
where $\mu_G$, $\mu_R$ are mean vectors and $\Sigma_G$, $\Sigma_R$ are full covariance matrices of activations for molecules from sets $G$ and $R$ respectively. FCD correlates with other metrics. For example, if the generated structures are not diverse enough (low $\textrm{IntDiv}_p$) or the model produces too many duplicates (low uniqueness), FCD will decrease, since the variance is smaller. We suggest using FCD for hyperparameter tuning and final model selection. Values of this metric are non-negative, lower is better.

{\bf Properties distribution} is a useful tool for visually assessing the generated structures. To quantitatively compare the distributions in the generated and test sets, we compute a 1D Wasserstein-1 distance between property distributions of generated and test sets. We also visualize a kernel density estimation of these distributions in the Experiments section. We use the following four properties:
\begin{itemize}
\item {\bf Molecular weight (MW)}: the sum of atomic weights in a molecule. By plotting histograms of molecular weight for the generated and test sets, one can judge if a generated set is biased towards lighter or heavier molecules.

\item {\bf LogP}: the octanol-water partition coefficient, a ratio of a chemical's concentration in the octanol phase to its concentration in the aqueous phase of a two-phase octanol/water system; computed with RDKit’s Crippen \citep{Wildman1999-fp} estimation.

\item {\bf Synthetic Accessibility Score (SA)}: a heuristic estimate of how hard $(10)$ or how easy $(1)$ it is to synthesize a given molecule. SA score is based on a combination of the molecule's fragments contributions \citep{Ertl2009-vd}. Note that SA score does not adequately assess up-to-date chemical structures, but it is useful for assessing distribution learning models.

\item {\bf Quantitative Estimation of Drug-likeness (QED)}: a $[0, 1]$ value estimating how likely a molecule is a viable candidate for a drug. QED is meant to capture the abstract notion of aesthetics in medicinal chemistry \citep{Bickerton2012-sd}. Similar to SA, descriptor limits in QED have been changing during the last decade and current limits may not cover latest drugs \citep{shultz2018two}.

\end{itemize}

\section{Dataset} \label{sec:dataset}
The proposed dataset used for training and testing is based on the ZINC Clean Leads \citep{Sterling2015-gf} collection which contains $4{,}591{,}276$ molecules with molecular weight in the range from $250$ to $350$ Daltons, a number of rotatable bonds not greater than $7$, and XlogP \citep{wang1997new} not greater then $3.5$. Clean-leads dataset consists of structures suitable for identifying hit compounds and they are small enough to allow for further ADMET optimization of generated molecules \citep{teague1999design}. We removed molecules containing charged atoms, atoms besides C, N, S, O, F, Cl, Br, H, or cycles larger than $8$ atoms. The molecules were filtered via custom medicinal chemistry filters (MCFs) and PAINS filters \citep{Baell2010-li}. We describe MCFs and discuss PAINS in Appendix~\ref{ap1}. We removed charged molecules to avoid ambiguity with tautomers and pH conditions. Note that in the initial set of molecules, functional groups were present in both ionized and unionized forms.

The final dataset contains $1{,}936{,}963$ molecules, with internal diversity $\textrm{IntDiv}_1=0.857$; it contains $448{,}854$ unique Bemis-Murcko \citep{Bemis1996-js} scaffolds and $58{,}315$ unique BRICS \citep{Degen2008-fk} fragments. We show example molecules in Figure~\ref{fig:diverse} and a representative diverse subset in Appendix~\ref{ap2}. We provide recommended split into three non-intersecting parts: train ($1{,}584{,}664$ molecules), test ($176{,}075$ molecules) and scaffold test ($176{,}226$ molecules). The scaffold test set has all molecules containing a Bemis-Murcko scaffold from a random subset of scaffolds. Hence, scaffolds from the scaffold test set differ from scaffolds in both train and test sets. We use scaffold test split to assess whether a model can produce novel scaffolds absent in the training set. The test set is a random subset of the remaining molecules in the dataset.

\begin{figure}[t]
        \centerline{
        \includegraphics[width=0.3\columnwidth]{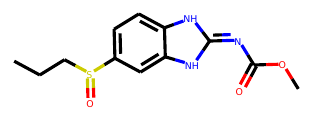}~\includegraphics[width=0.22\columnwidth]{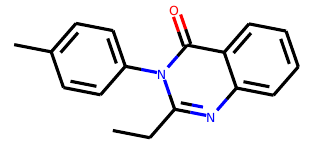}~\includegraphics[width=0.3\columnwidth]{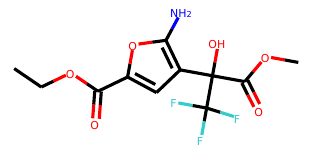}}

        \centerline{\includegraphics[width=0.24\columnwidth]{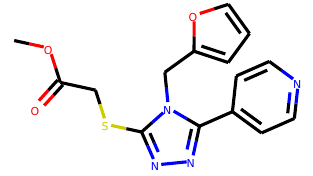}~\includegraphics[width=0.3\columnwidth]{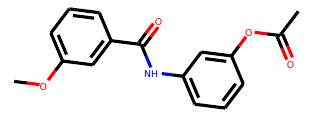}~\includegraphics[width=0.3\columnwidth]{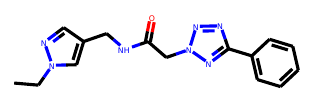}}

        \caption{Examples of molecules from MOSES dataset.}
        \label{fig:diverse}
\end{figure}

\section{Baselines}
We implemented several models that cover different approaches to molecular generation, such as character-level recurrent neural networks (CharRNN) \citep{Segler2018-eo, Preuer2018-pf}, Variational Autoencoders (VAE) \citep{Gomez-Bombarelli2018-jp, Kadurin2017-rv, Blaschke2018-oa}, Adversarial Autoencoders (AAE) \citep{Kadurin2017-rv, Polykovskiy2018-rf}, Junction Tree Variational Autoencoders (JTN-VAE) \citep{Jin2018-sm}, LatentGAN \citep{prykhodko2019novo}, and non-neural baselines.

Model comparison can be challenging since different training parameters (number of epochs, batch size, learning rate, initial state, optimizer) and architecture hyperparameters (hidden layer dimension, number of layers, etc.) can significantly alter their performance. For each model, we attempted to preserve its original architecture as published and tuned the hyperparameters to improve the performance. We used random search over multiple architectures for every model and selected the architecture that produced the best value of FCD. Models are implemented in Python $3$ utilizing PyTorch \citep{Paszke2017-lk} framework. Please refer to the Appendix~\ref{ap3} for the training details and hyperparameters. 

{\bf Character-level recurrent neural network (CharRNN)} \citep{Segler2018-eo} models a distribution over the next token given previously generated ones. We train this model by maximizing log-likelihood of the training data represented as SMILES strings.

{\bf Variational autoencoder (VAE)} \citep{Kingma2013} consists of two neural networks---an encoder and a decoder---that infer a mapping from high-dimensional data representation onto a lower-dimensional space and back. The lower-dimensional space is called the latent space, which is often a continuous vector space with normal prior distribution. VAE parameters are optimized to encode and decode data by minimizing reconstruction loss and regularization term in a form of Kullback-Leibler divergence. VAE-based architecture for the molecular generation was studied in multiple previous works \citet{Kadurin2017-rv, Blaschke2018-oa, Gomez-Bombarelli2018-jp}. We combine aspects from these implementations and use SMILES as input and output representations.

{\bf Adversarial Autoencoder (AAE)} \citep{Makhzani2015} replaces the Kullback-Leibler divergence from VAE with an adversarial objective. An auxiliary discriminator network is trained to distinguish samples from a prior distribution and model's latent codes. The encoder then adapts its latent codes to minimize discriminator's predictive accuracy. The training process oscillates between training the encoder-decoder pair and the discriminator. Unlike Kullback-Leibler divergence that has a closed-form analytical solution only for a handful of distributions, a discriminator can be used for any prior distribution. AAE-based models for molecular design were studied in \citep{Kadurin2017-ki, Kadurin2017-rv,Polykovskiy2018-rf}. Similar to VAE, we use SMILES as input and output representations.

{\bf Junction Tree VAE (JTN-VAE)} \citep{Jin2018-sm} generates molecules in two phases by exploiting valid subgraphs as components. In the first phase, it generates a tree-structured object (a junction tree) whose role is to represent the scaffold of subgraph components and their coarse relative arrangements. The components are valid chemical substructures automatically extracted from the training set. In the second phase, the subgraphs (nodes of the tree) are assembled together into a coherent molecular graph.

{\bf Latent Vector Based Generative Adversarial Network (LatentGAN)} \citep{prykhodko2019novo} combines an autoencoder and a generative adversarial network. LatentGAN pretrains an autoencoder to map SMILES structures onto latent vectors. A generative adversarial network is then trained to produce latent vectors for the pre-trained decoder.

{\bf Non-neural baselines} implemented in MOSES are n-gram generative model, Hidden Markov Model (HMM), and a combinatorial generator. N-gram model collects statistics of n-grams frequencies in the training set and uses such distribution to sequentially sample new strings. Hidden Markov models utilize Baum-Welch algorithm to learn a probabilistic distribution over the SMILES strings. The model consists of several states ($s_1, \dots, s_K$), transition probabilities between states $p(s_{i+1} \mid s_i)$, and token emission probabilities $p(x_i \mid s_i)$. Beginning from a "start" state, at each iteration the model samples a next token and state from emission and transition probabilities correspondingly. A combinatorial generator splits molecular graphs of the training data into BRICS fragments and generates new molecules by randomly connecting random substructures. We sample fragments according to their frequencies in the training set to model the distribution better.

\section{Platform}
The dataset, metrics and baseline models are provided in a GitHub repository \url{https://github.com/molecularsets/moses} and as a PyPI package {\tt molsets}. To contribute a new model, one should train a model on MOSES train set, generate $30{,}000$ samples and compute metrics using the provided utilities. We recommend running the experiment at least three times with different random seeds to estimate sensitivity of the model to random parameter initialization. We store molecular structures in SMILES format; molecular graphs can be reconstructed using RDKit \citep{Landrum_undated-rr}.

\section{Results}
We trained the baseline models on MOSES train set and provide results in this section. In Table~\ref{tab:results1} we compare models with respect to the validity and uniqueness metrics. Hidden Markov Model and NGram models fail to produce valid molecules since they have a limited context. Combinatorial generator and JTN-VAE have built-in validity constraints, so their validity is 100\%.

\begin{table}[t]
\caption{Performance metrics for baseline models: fraction of valid molecules, fraction of unique molecules from $1{,}000$ and $10{,}000$ molecules. Reported (mean $\pm$ std) over three independent model initializations.}
\label{tab:results1}
\centering
\addtolength{\leftskip} {-4cm}
\addtolength{\rightskip}{-4cm}

\begin{tabular}{llllllll}
\toprule
Model  & Valid ($\uparrow$)  & Unique@1k ($\uparrow$)  & Unique@10k ($\uparrow$) \\
\midrule
 {\it Train } 
 & {\it 1.0 } 
 & {\it 1.0 } 
 & {\it 1.0} \\ \midrule
 HMM
 & 0.076 {\tiny $\pm$ 0.0322}
 & 0.623 {\tiny $\pm$ 0.1224}
 & 0.5671 {\tiny $\pm$ 0.1424}  \\
NGram
 & 0.2376 {\tiny $\pm$ 0.0025}
 & 0.974 {\tiny $\pm$ 0.0108}
 & 0.9217 {\tiny $\pm$ 0.0019}  \\
Combinatorial
 & {\bf 1.0 {\tiny $\pm$ 0.0}}
 & 0.9983 {\tiny $\pm$ 0.0015}
 & 0.9909 {\tiny $\pm$ 0.0009} \\
CharRNN
 & 0.975 {\tiny $\pm$ 0.026}
 & {\bf 1.0  {\tiny $\pm$ 0.0}}
 & 0.999 {\tiny $\pm$ 0.0}
 \\ 
VAE
 & 0.977 {\tiny $\pm$ 0.001}
 & {\bf 1.0 {\tiny $\pm$ 0.0} }
 & 0.998 {\tiny $\pm$ 0.001}
\\ 
AAE
 & 0.937 {\tiny $\pm$ 0.034}
 & {\bf 1.0 {\tiny $\pm$ 0.0} }
 & 0.997 {\tiny $\pm$ 0.002}
 \\ 
JTN-VAE
 & {\bf 1.0  {\tiny $\pm$ 0.0} }
 & {\bf 1.0  {\tiny $\pm$ 0.0} }
 & {\bf 0.9996 {\tiny $\pm$0.0003} } \\
LatentGAN
 & 0.897 {\tiny $\pm$ 0.002}
 & {\bf 1.0 {\tiny $\pm$ 0.0} }
 & 0.997 {\tiny $\pm$ 0.005}
 \\
\bottomrule
\end{tabular}
\end{table}

Table~\ref{tab:results2} reports additional properties of the generated set: fraction of molecules passing filters, fraction of molecules not present in the training set, and internal diversity. All modules successfully avoid forbidden structures (MCF and PAINS) even though such restrictions were only defined implicitly---using a training dataset. Combinatorial generator has higher diversity than the training dataset, which might be favorable for discovering new chemical structures. Autoencoder-based models show low novelty, indicating that these models overfit to the training set.

\begin{table}[t]
\caption{Performance metrics for baseline models: fraction of molecules passing filters (MCF, PAINS, ring sizes, charge, atom types), novelty, and internal diversity. Reported (mean $\pm$ std) over three independent model initializations.}
\label{tab:results2}
\centering
\addtolength{\leftskip} {-4cm}
\addtolength{\rightskip}{-4cm}

\begin{tabular}{llllllll}
\toprule
Model  & Filters ($\uparrow$) & Novelty ($\uparrow$) & IntDiv$_1$ & IntDiv$_2$\\
\midrule
 {\it Train } 
 & {\it 1.0 } 
 & {\it 0.0} 
 & {\it 0.857 } 
 & {\it 0.851 } 
\\ \midrule
 HMM
 & 0.9024 {\tiny $\pm$ 0.0489}
 & {\bf 0.9994 {\tiny $\pm$ 0.001}} 
 & 0.8466 {\tiny $\pm$ 0.0403}
 & 0.8104  {\tiny $\pm$ 0.0507} 
\\
NGram
 & 0.9582 {\tiny $\pm$ 0.001}
 & 0.9694 {\tiny $\pm$ 0.001} 
 & {\bf 0.8738 {\tiny $\pm$ 0.0002} }
 & 0.8644 {\tiny $\pm$ 0.0002} 
\\
Combinatorial
 & 0.9557 {\tiny $\pm$ 0.0018}
 & 0.9878 {\tiny $\pm$ 0.0008} 
 & 0.8732 {\tiny $\pm$ 0.0002}
 & {\bf 0.8666 {\tiny $\pm$ 0.0002} } 
\\
CharRNN
 & 0.994 {\tiny $\pm$ 0.003}
 & 0.842 {\tiny $\pm$ 0.051}
 & 0.856 {\tiny $\pm$ 0.0}
 & 0.85 {\tiny $\pm$ 0.0}

 \\ 
VAE
 & {\bf 0.997 {\tiny $\pm$ 0.0} }
 & 0.695 {\tiny $\pm$ 0.007}
 & 0.856 {\tiny $\pm$ 0.0}
 & 0.85 {\tiny $\pm$ 0.0 }

\\ 
AAE
 & 0.996 {\tiny $\pm$ 0.001}
 & 0.793 {\tiny $\pm$ 0.028}
 & 0.856 {\tiny $\pm$ 0.003}
 & 0.85 {\tiny $\pm$ 0.003}

 \\ 
JTN-VAE
 & 0.976 {\tiny $\pm$ 0.0016}
 & 0.9143 {\tiny $\pm$ 0.0058}
 & 0.8551 {\tiny $\pm$ 0.0034}
 & 0.8493 {\tiny $\pm$ 0.0035}
\\
LatentGAN
 & 0.973 {\tiny $\pm$ 0.001}
 & 0.949 {\tiny $\pm$ 0.001} 
 & 0.857 {\tiny $\pm$ 0.0}
 & 0.85 {\tiny $\pm$ 0.0 }
 \\
\bottomrule
\end{tabular}
\end{table}

Table~\ref{tab:results3} reports Fr\'echet ChemNet Distance (FCD) and similarity to a nearest neighbor (SNN). All neural network-based models show low FCD, indicating that the models successfully captured the statistics of the dataset. Surprisingly, a simple language model, character level RNN, shows the best results in terms of the FCD measure. Variational autoencoder (VAE) showed the best results in terms of SNN, but combined with low novelty we suppose that the model overfitted on the training set.

\begin{table}[t]
\caption{Performance metrics for baseline models: Fr\'echet ChemNet Distance (FCD) and Similarity to a nearest neighbor (SNN); Reported (mean $\pm$ std) over three independent model initializations. Results for random test set (Test) and scaffold split test set (TestSF).}
\label{tab:results3}
\centering
\addtolength{\leftskip} {-4cm}
\addtolength{\rightskip}{-4cm}

\begin{tabular}{llllll}
\toprule
\multirow{2}{*}{Model}  &  \multicolumn{2}{c}{FCD ($\downarrow$)} & \multicolumn{2}{c}{SNN ($\uparrow$)} \\
\cmidrule(l){2-3} \cmidrule(r){4-5}
&  Test & TestSF &  Test & TestSF \\
\midrule
 {\it Train } 
 & {\it 0.008 }
 & {\it 0.476 }
 & {\it 0.642 }
 & {\it 0.586 }
\\ \midrule
 HMM
 & 24.4661 {\tiny $\pm$ 2.5251}
 & 25.4312 {\tiny $\pm$ 2.5599}
 & 0.3876 {\tiny $\pm$ 0.0107}
 & 0.3795 {\tiny $\pm$ 0.0107}
\\
NGram
 & 5.5069 {\tiny $\pm$ 0.1027}
 & 6.2306 {\tiny $\pm$ 0.0966}
 & 0.5209 {\tiny $\pm$ 0.001}
 & 0.4997 {\tiny $\pm$ 0.0005}
\\
Combinatorial
 & 4.2375 {\tiny $\pm$ 0.037}
 & 4.5113 {\tiny $\pm$ 0.0274}
 & 0.4514 {\tiny $\pm$ 0.0003}
 & 0.4388 {\tiny $\pm$ 0.0002}
\\
CharRNN
 & {\bf 0.073 {\tiny $\pm$ 0.025}}
 & {\bf 0.52 {\tiny $\pm$ 0.038}}
 & 0.601 {\tiny $\pm$ 0.021}
 & 0.565 {\tiny $\pm$ 0.014}

 \\ 
VAE
 & 0.099 {\tiny $\pm$ 0.013 }
 & 0.567 {\tiny $\pm$ 0.034 }
 & {\bf 0.626 {\tiny $\pm$ 0.0 }}
 & {\bf 0.578 {\tiny $\pm$ 0.001 } }

\\ 
AAE
 & 0.556 {\tiny $\pm$ 0.203}
 & 1.057 {\tiny $\pm$ 0.237}
 & 0.608 {\tiny $\pm$ 0.004}
 & 0.568 {\tiny $\pm$ 0.005}
 \\ 
JTN-VAE
 & 0.3954 {\tiny $\pm$ 0.0234}
 & 0.9382 {\tiny $\pm$ 0.0531}
 & 0.5477 {\tiny $\pm$ 0.0076}
 & 0.5194 {\tiny $\pm$ 0.007}
\\
LatentGAN
 & 0.296 {\tiny $\pm$ 0.021}
 & 0.824 {\tiny $\pm$ 0.030}
 & 0.538 {\tiny $\pm$ 0.001}
 & 0.514 {\tiny $\pm$ 0.009}
 \\
\bottomrule
\end{tabular}
\end{table}

In Table~\ref{tab:results4} we report similarities of substructure distributions---fragments and scaffolds. Scaffold similarity from the training set to the scaffold test set (TestSF) is zero by design. Note that CharRNN successfully discovered many novel scaffolds (11\%), suggesting that the model generalizes well.

\begin{table*}[t]
\caption{Fragment similarity (Frag), Scaffold similarity (Scaff). Reported (mean $\pm$ std) over three independent model initializations. Results for random test set (Test) and scaffold split test set (TestSF).}
\label{tab:results4}
\centering
\addtolength{\leftskip} {-4cm}
\addtolength{\rightskip}{-4cm}

\begin{tabular}{llllllllll}
\toprule
\multirow{2}{*}{Model} & \multicolumn{2}{c}{Frag ($\uparrow$)} & \multicolumn{2}{c}{Scaf ($\uparrow$)}  \\
\cmidrule(r){2-3} \cmidrule(lr){4-5} \cmidrule(lr){6-7} \cmidrule(l){8-9} 
& Test & TestSF & Test & TestSF \\
\midrule
 {\it Train } & {\it 1.0 }
 & {\it 0.999 }
 & {\it 0.991 }
 & {\it 0.0 }\\ \midrule
HMM
 & 0.5754 {\tiny $\pm$ 0.1224}
 & 0.5681 {\tiny $\pm$ 0.1218}
 & 0.2065 {\tiny $\pm$ 0.0481}
 & 0.049 {\tiny $\pm$ 0.018} \\
NGram
 & 0.9846 {\tiny $\pm$ 0.0012}
 & 0.9815 {\tiny $\pm$ 0.0012}
 & 0.5302 {\tiny $\pm$ 0.0163}
 & 0.0977 {\tiny $\pm$ 0.0142} \\
Combinatorial
 & 0.9912 {\tiny $\pm$ 0.0004}
 & 0.9904 {\tiny $\pm$ 0.0003}
 & 0.4445 {\tiny $\pm$ 0.0056}
 & 0.0865 {\tiny $\pm$ 0.0027} \\
CharRNN
 & {\bf 1.0 {\tiny $\pm$ 0.0} }
 & {\bf 0.998 {\tiny $\pm$ 0.0} }
 & 0.924 {\tiny $\pm$ 0.006}
 & {\bf 0.11 {\tiny $\pm$ 0.008} } \\
VAE
 & 0.999 {\tiny $\pm$ 0.0 }
 & {\bf 0.998 {\tiny $\pm$ 0.0 } }
 & {\bf 0.939 {\tiny $\pm$ 0.002 } }
 & 0.059{\tiny $\pm$ 0.01 } \\
AAE
 & 0.991 {\tiny $\pm$ 0.005}
 & 0.99 {\tiny $\pm$ 0.004}
 & 0.902 {\tiny $\pm$ 0.037}
 & 0.079 {\tiny $\pm$ 0.009} \\
JTN-VAE
 & 0.9965 {\tiny $\pm$ 0.0003}
 & 0.9947 {\tiny $\pm$ 0.0002}
 & 0.8964 {\tiny $\pm$ 0.0039}
 & 0.1009 {\tiny $\pm$ 0.0105} \\
LatentGAN
 & 0.999 {\tiny $\pm$ 0.003}
 & {\bf 0.998 {\tiny $\pm$ 0.003}}
 & 0.886 {\tiny $\pm$ 0.015}
 & 0.1 {\tiny $\pm$ 0.006} \\
\bottomrule
\end{tabular}
\end{table*}

\begin{figure}[t]
        \centerline{
        \includegraphics[width=0.5\columnwidth]{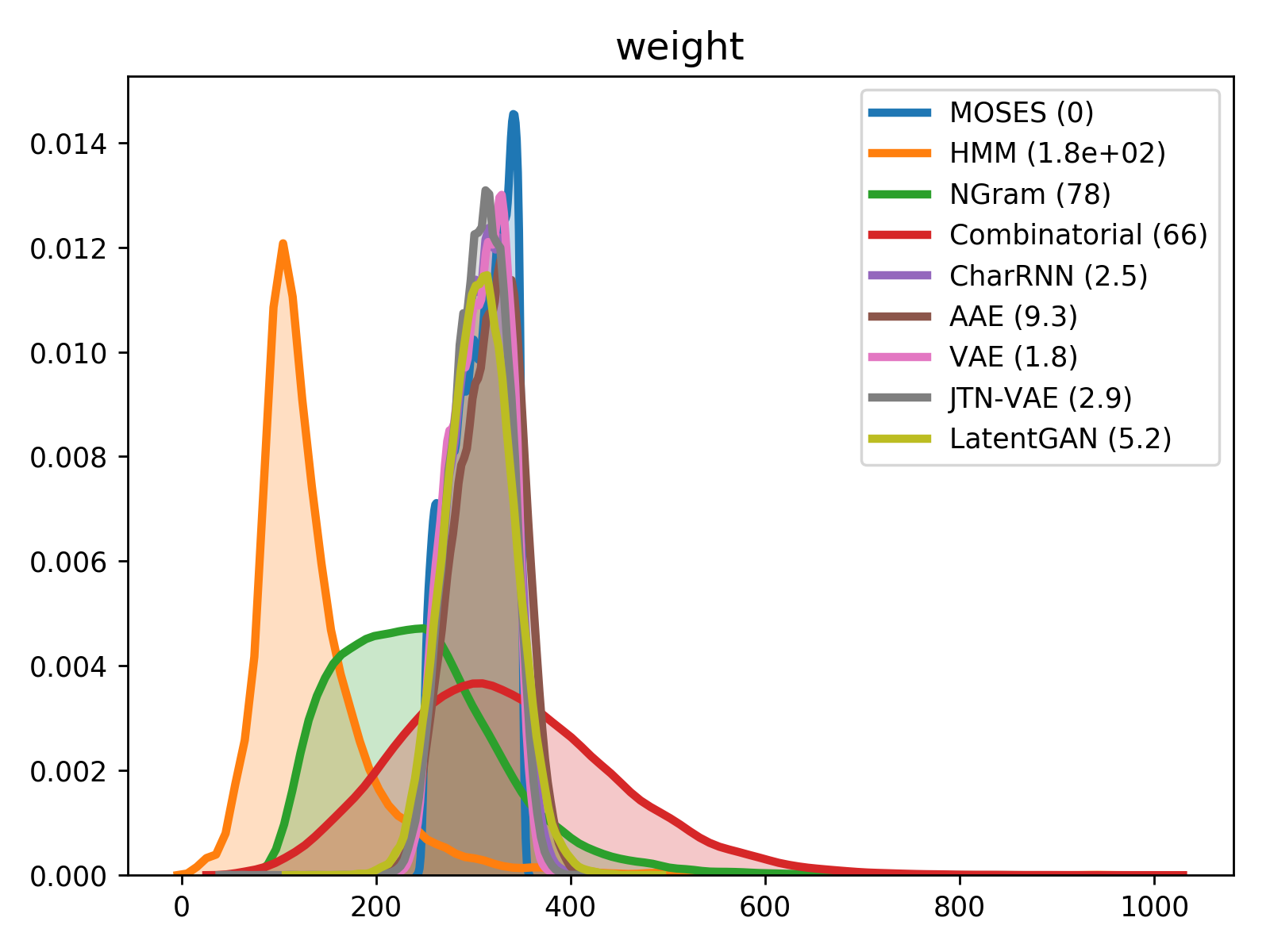}~\includegraphics[width=0.5\columnwidth]{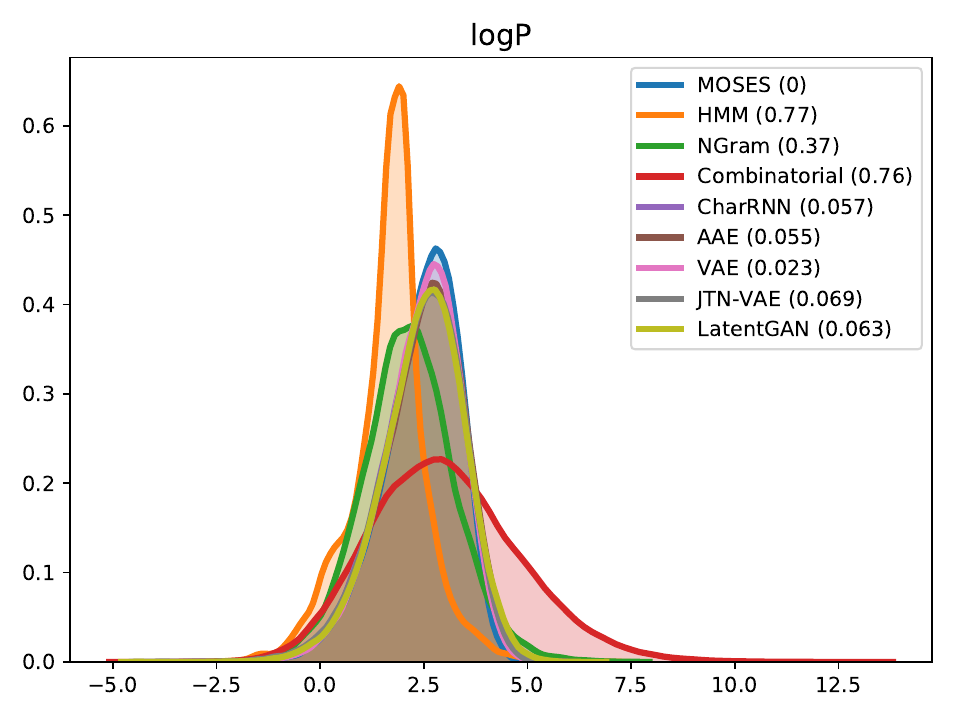}}

        \centerline{\includegraphics[width=0.5\columnwidth]{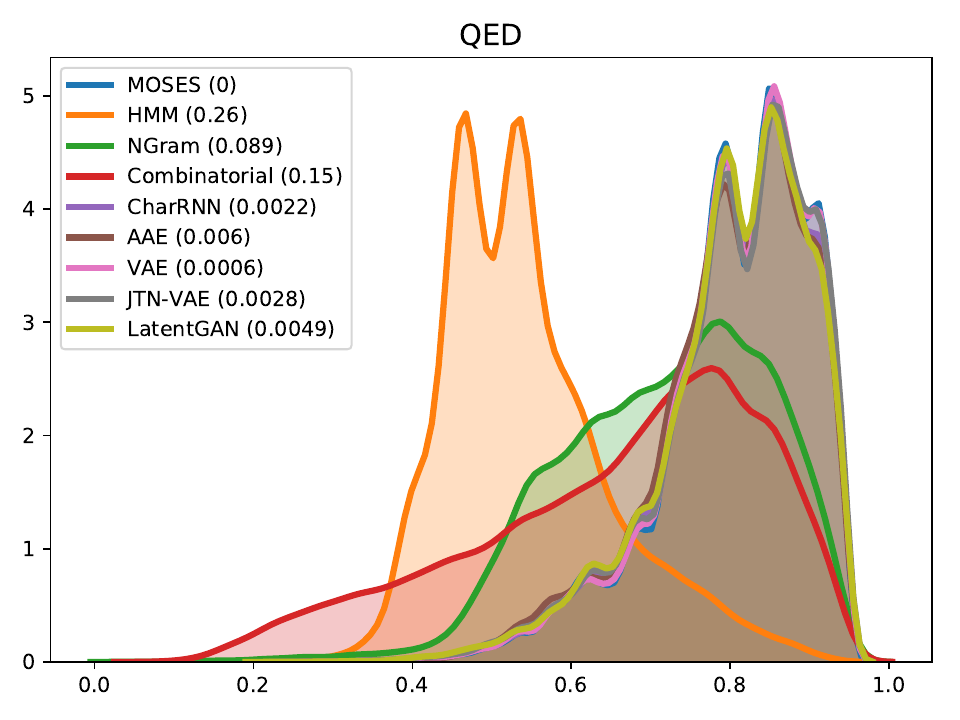}~\includegraphics[width=0.5\columnwidth]{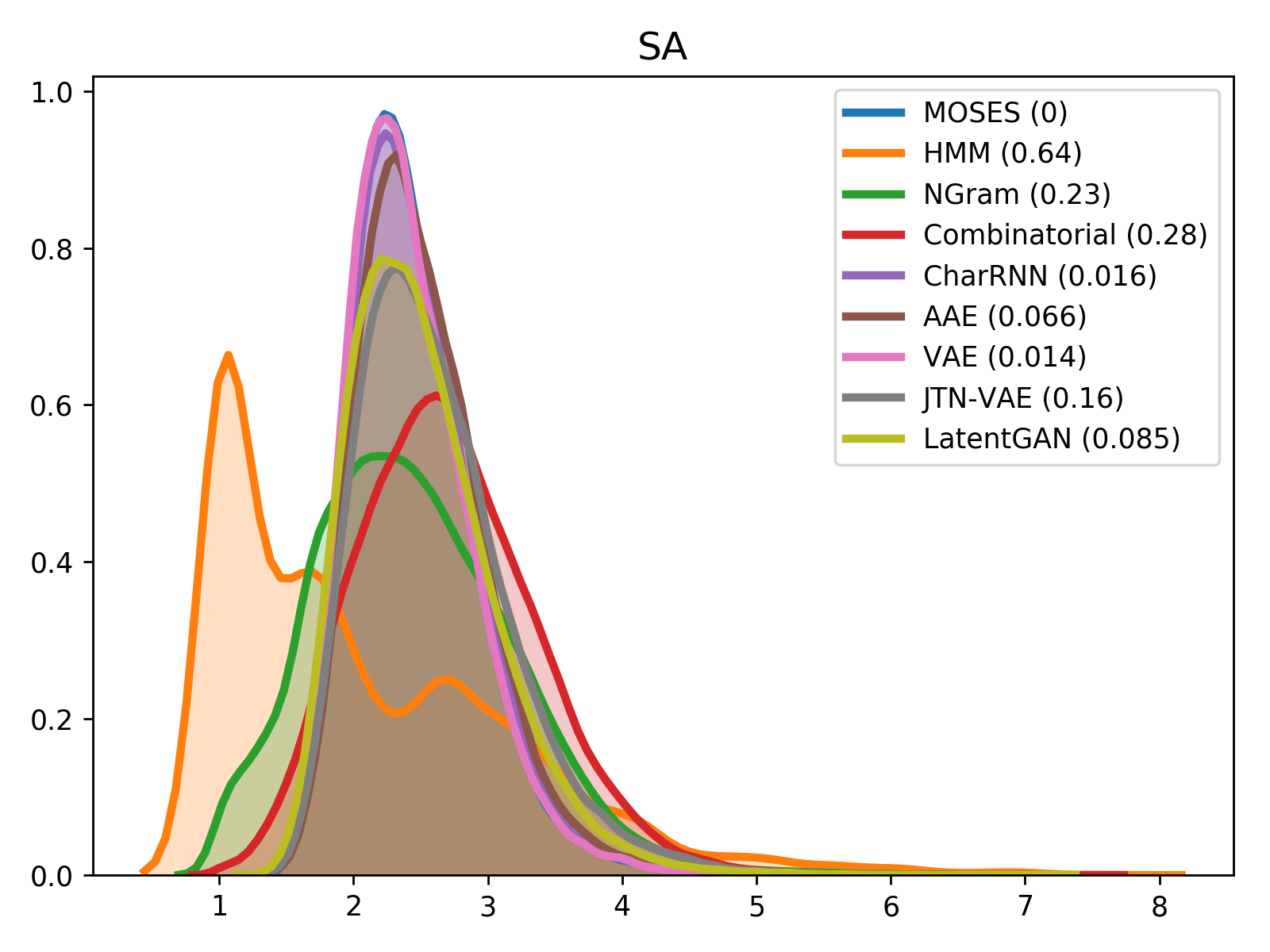}}

        \caption{Distribution of chemical properties for MOSES dataset and sets of generated molecules. In brackets---Wasserstein-1 distance to MOSES test set. Parameters: molecular weight, octanol-water partition coefficient (logP), quantitative estimation of drug-likeness (QED) and synthetic accessibility score (SA).}
        \label{fig:results}
\end{figure}

Finally, we compared distributions of four molecular properties in generated and test sets (Figure~\ref{fig:results}): molecular weight (MW), octanol-water partition coefficient (logP), quantitative estimation of drug-likeness (QED), and synthetic accessibility score (SA).  Deep generative models closely match the data distribution; hidden Markov Model is biased towards lighter molecules, which is consistent with low validity: larger molecules impose more validity constraints. A combinatorial generator has higher variance in molecular weight, producing larger and smaller molecules than those present in the training set.

\section{Discussion}
From a wide range of presented models, CharRNN currently performs the best in terms of the key metrics. Specifically, it produces the best FCD, Fragment, and Scaffold scores, indicating that the model not only captured the training distribution well, but also did not overfit on the training set.

The presented set of metrics assesses models' performance from different perspectives; therefore, for each specific downstream task, one could consider the most relevant metric. For example, evaluation based on Scaf/TestSF score could be relevant when model's objective is to discover novel scaffolds. For a general evaluation, we suggest using FCD/Test metric that captures multiple aspects of other metrics in a single number. However, it does not give insights into specific issues that cause high FCD/Test values, hence more interpretable metrics presented in this paper are necessary to investigate the model’s performance thoroughly.

\section{Conclusion}
With MOSES, we have designed a molecular generation benchmark platform that provides a dataset with molecular structures, an implementation of baseline models, and metrics for their evaluation. While standardized comparative studies and test sets are essential for the progress of machine learning applications, the current field of de-novo drug design lacks evaluation protocols for generative machine learning models. Being on the intersection of mathematics, computer science, and chemistry, these applications are often too challenging to explore for research scientists starting in the field. Hence, it is necessary to develop a transparent approach to implementing new models and assessing their performance. We presented a benchmark suite with unified and extendable programming interfaces for generative models and evaluation metrics.

This platform should allow for a fair and comprehensive comparison of new generative models. For future work on this project, we will keep extending the MOSES repository with new baseline models and new evaluation metrics. We hope this work will attract researchers interested in tackling drug discovery challenges.

\section*{Correspondence}
Correspondence to: alex@insilico.com, alan@aspuru.com, snikolenko@gmail.com.

\bibliographystyle{unsrtnat}

\appendix

\section{Medicinal chemistry filters and PAINS filters} \label{ap1}
Medicinal chemistry filters are used to discard compounds containing so-called ``structural alerts''. Molecules containing such moieties either bear unstable or reactive groups or undergo biotransformations resulting in the formation of toxic metabolites or intermediates.

We filtered the dataset with medicinal chemistry filters (MCFs) that we explain in this section. We used MCFs for rational pre-selection of compounds more appropriate for modern drug design and development. These include some electrophilic alkylating groups, such as Michael acceptors (MCF1-3), alkyl halides (MCF4), epoxide (MCF5), isocyanate (MCF6), aldehyde (MCF7), imine (Schiff base, MCF8), aziridine (MCF9) which are very liable for nucleophilic attack. In many cases, it leads to unselective protein and/or DNA damage. Metabolism of hydrazine (MCF10) furnishes diazene intermediates (MCF11), which are also alkylating warheads. Monosubstituted furans (MCF12) and thiophenes (MCF13) are transformed into reactive intermediates via epoxidation. Their active metabolites irreversibly bind nucleophilic groups and modify proteins. Electrophilic aromatics (e.g. halopyridine, MCF14), oxidized anilines (MCF15) and disulfides (MCF16) are also highly reactive. In vivo, alkylators are trapped and inactivated by the thiol group of glutathione, which is a key natural antioxidant. Azides (MCF17) are highly toxic; compounds containing this functional group particularly cause genotoxicity. Aminals (MCF18) and acetals (MCF19) are frequently unstable and inappropriate in generated structures. In addition, virtual structures containing a large number of halogens (MCF20-22) should be excluded due to increased molecular weight and lipophilicity (insufficient solubility for oral administration), metabolic stability, and toxicity. The detailed mechanism of toxicity for structure alerts mentioned above has been comprehensively described in \citep{Kalgutkar2005-hj, Kalgutkar2005-th}.

PAINS (pan-assay interfering compounds) filters are the set of substructure filters proposed to use for reducing the number of false positives, assay artifacts and unspecific bioactive molecules in the screening libraries. It was stated that the presence of certain fragments in a structure could lead to undesirable properties (reactivity, chelation, the formation of colloidal aggregates, dyes) affecting assay results. It should be noted that the analysis of available data from the PubChem database clearly demonstrated the limitations of PAINS filters \citep{capuzzi2017phantom, senger2016filtering}. Indeed, PAINS were observed among the molecules inactive in at least 100 bioassays (the dark chemical matter). Interestingly, structural analysis of well-known drugs revealed PAINS among them. For instance, quinone-based compounds were classified as PAINS, however there are quinone-based drugs approved by the FDA in the market. Despite mentioned above, this approach can be considered as a viable tool for narrowing down the large virtual chemical spaces produced by generative models to drug-like chemical matter.

\section{Diverse set of molecules from MOSES} \label{ap2}
\begin{figure}[hb]
        \begin{center}
        \centerline{\includegraphics[width=1\textwidth]{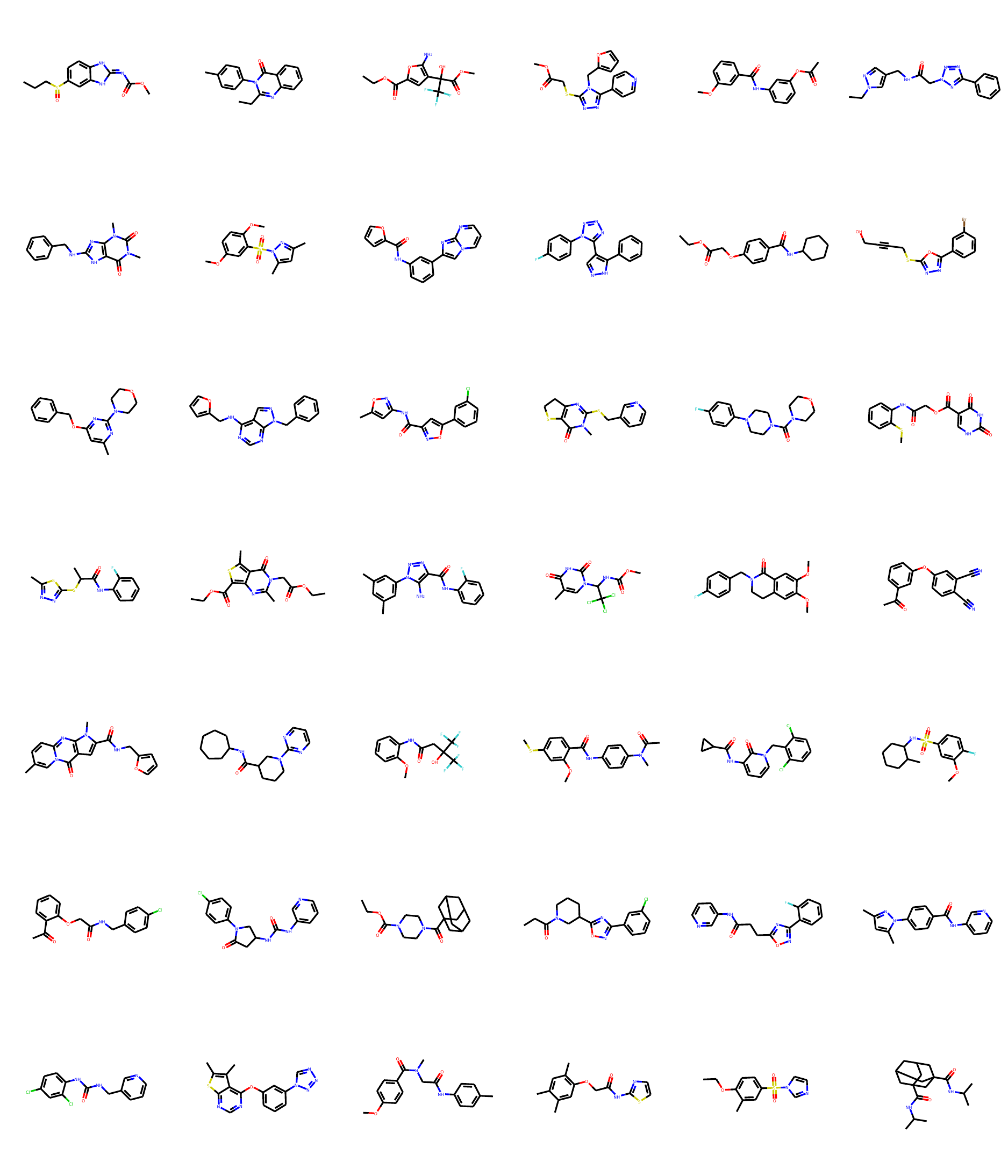}}
        \caption{A diverse subset of molecules from MOSES dataset.}
        \label{fig:diverse2}
        \end{center}
\end{figure}

In Figure~\ref{fig:diverse2}, we show a diverse set of molecules of MOSES dataset. We obtained these molecules by iteratively adding structures with the lowest cosine similarity to the nearest compound in the currently selected set.

\section{Hyperparameters and training details} \label{ap3}
 
\begin{figure}[ht]
        \begin{center}
        \centerline{\includegraphics[width=0.5\columnwidth]{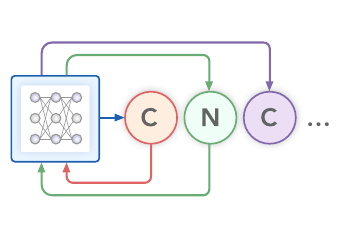}}
        \caption{CharRNN model. A model is trained by maximizing the likelihood of known molecules.}
        \label{fig:charrnn}
        \end{center}
\end{figure}

{\bf Character-level recurrent neural networks (CharRNN)} used Long Short-Term Memory \citep{Hochreiter1997-mv} cells stacked into $3$ layers with hidden dimension $768$ each. We used a dropout \citep{Srivastava2013-oh} layer with dropout rate $0.2$. Softmax was utilized as an output layer. Training was done with a batch size of $64$, using the Adam \citep{Kingma2015-ot} optimizer for $80$ epochs with a learning rate of $10^{-3}$ that halved after each $10$ epochs. We display CharRNN model in Figure~\ref{fig:charrnn}.

\begin{figure}[ht]
        \begin{center}
        \centerline{\includegraphics[width=1\columnwidth]{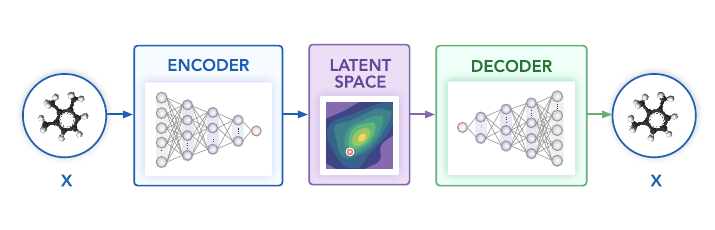}}
        \caption{Autoencoder-based models. VAE/AAE forms a specific distribution in the latent space.}
        \label{fig:vae_aae}
        \end{center}
\end{figure}

{\bf Variational autoencoder (VAE)} used a bidirectional \citep{Schuster1997-dc} Gated Recurrent Unit (GRU) \citep{Cho2014-wf} with a linear output layer as an encoder. The decoder was a $3$-layer GRU of $512$ hidden dimensions with intermediate dropout layers with dropout probability $0.2$. Training was done with a batch size of $128$, utilizing a gradient clipping of $50$, KL-term weight linearly increased from $0$ to $1$ during training. We optimized the model using Adam optimizer with a learning rate of $3 \cdot 10^{-4}$. We trained the model for $100$ epochs. We display an autoencoder model in Figure~\ref{fig:vae_aae}.

{\bf Adversarial Autoencoders (AAE)} consisted of an encoder with a single layer bidirectional LSTM with $512$ hidden dimensions, a decoder with a $2$-layer LSTM with $512$ hidden dimensions and a shared embedding of size $128$. The discriminator network was a 2-layer fully connected neural network with $640$ and $256$ nodes respectively with exponential linear unit (ELU) \citep{clevert2015fast} activation function \citep{Clevert2015-dz}. We trained a model with a batch size of $512$, with the Adam optimizer using a learning rate of $10^{-3}$ for $120$ epochs. We halved the learning rate after each $20$ epochs.

{\bf Junction Tree VAE (JT-VAE)} We report the experimental results from the official JT-VAE repository \citep{Jin2018-sm}.

\begin{figure}[ht]
        \begin{center}
        \centerline{\includegraphics[width=0.8\textwidth]{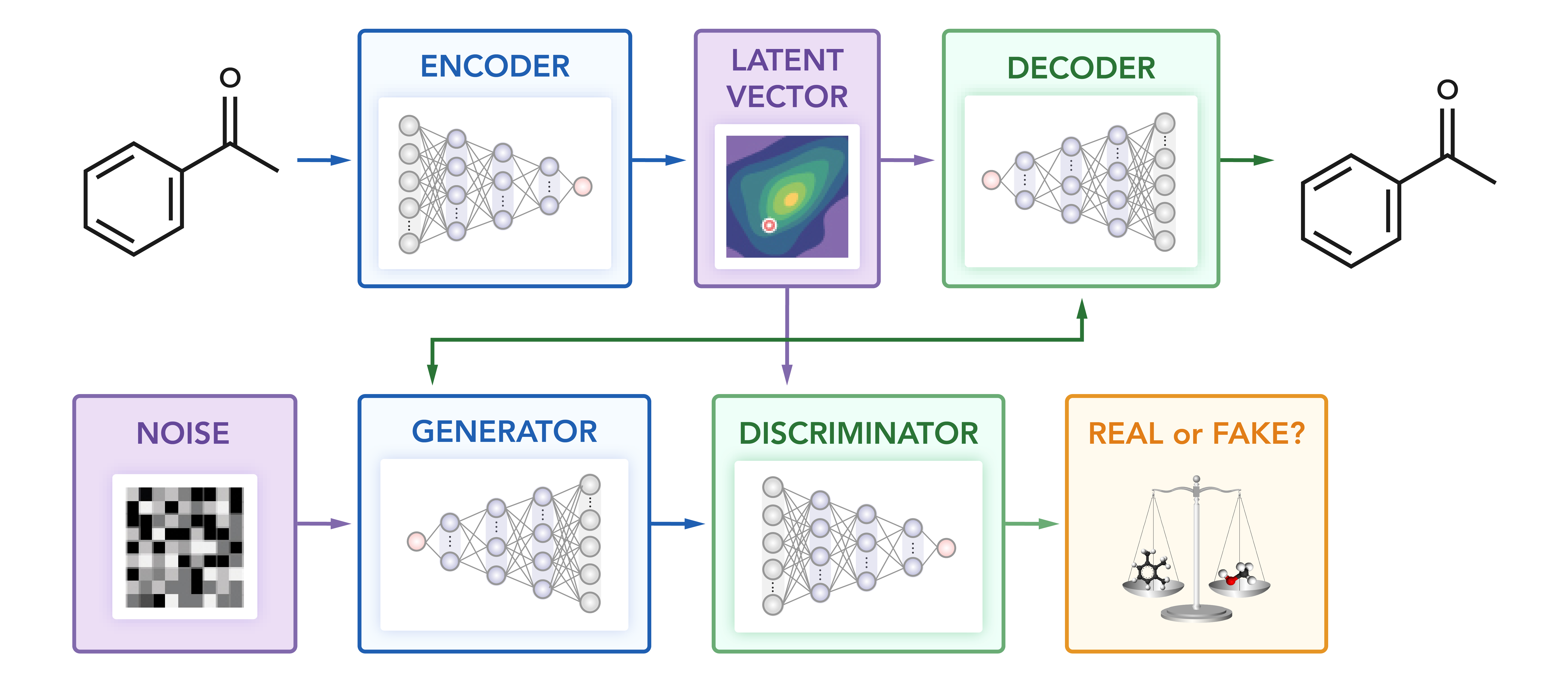}}
        \caption{LatentGAN model. A model combines an autoencoder and generative adversarial networks.}
        \label{fig:latentgan}
        \end{center}
\end{figure}

{\bf Latent Vector Based Generative Adversarial Network (LatentGAN)} pretrained an aused a heteroencoder \citep{bjerrum2018improving} containing a two-layer bidirectional encoder with $512$ LSTM units per layer. Authors added a Gaussian noise with a zero mean standard deviation of $0.1$ to the latent codes, resembling VAE with a fixed variance of proposal distributions. The LSTM decoder had $4$ layers. The neural network was trained on pairs of randomly chosen non-canonical SMILES strings \citep{bjerrum2017smiles}. The autoencoder network was trained for $100$ epochs with a batch size of $128$ sequences, using Adam optimizer with a learning rate $10^{-3}$ for first $50$ epochs and with an exponential learning rate decay reaching a value of $10^{-6}$ in the final epoch. LatentGAN uses Wasserstein GAN with gradient penalty (WGAN-GP) \citep{gulrajani2017improved} with a fully connected discriminator with $3$ layers of which the first two used the leaky ReLU activation function, and the last layer no activation function. The generator consisted of five fully connected layers with batch normalization and leaky ReLU activation. The GAN was trained for $2{,}000$ epochs using a learning rate of $2\cdot 10^{-4}$ with Adam parameters $\beta_1=0.5, \beta_2=0.9$. We display LatentGAN model in Figure~\ref{fig:latentgan}.

{\bf Combinatorial generator} randomly joins BRICS fragments. We first cut all molecules from the training set into fragments and compute the frequency of each fragment. We also compute a distribution of the number of fragments in the training set. To produce a molecule, we first randomly sample a total number of fragments that we will use in the molecule. We then iteratively sample fragments according to their frequencies. We omit fragments that will lead to invalid final molecules. For example, if there are currently two free attachment points in the molecule and two fragments left to attach, we cannot attach fragments with more than one attachment points. We also experimented with randomly sampling fragments until there are no more connection points. However, such method performed worse.

{\bf N-gram model} used 11-gram count statistics with pseudo-count of $0.01$. During generation, when there were no statistics available for the current (n-1)-gram, we reduced the context length until some statistics were available. In extreme cases, the context reduced to a single token which is equivalent to a bigram model.

{\bf Hidden Markov Model} uses Baum-Welch algorithm for training the model. We used HMM with $200$ states and trained the model for $100$ epochs on a subset of MOSES train set with first $100{,}000$ molecules. Note that HMM uses batch training which leads to high computational costs. To speedup learning, we used K-means$||$ algorithm \citep{bahmani2012scalable} to initialize parameters of the model.

\end{document}